\pdfoutput=1
\documentclass[11pt]{article}

\usepackage[table]{xcolor} 
\usepackage[final]{acl}

\usepackage{tabularx}%
\usepackage{multirow}
\usepackage{times}
\usepackage{latexsym}
\usepackage{booktabs}
\usepackage{arydshln}
\usepackage{amsmath}
\usepackage{bbding}
\usepackage{textcomp}
\usepackage{enumitem}

\usepackage{hyperref}
\usepackage[T1]{fontenc}
\usepackage[utf8]{inputenc}

\usepackage{microtype}

\usepackage{inconsolata}

\usepackage{graphicx}
\usepackage{multirow}
\usepackage{subfig}
\usepackage{color}
\usepackage{colortbl}
\usepackage{multirow}

\usepackage{listings,xfp}
\usepackage{anyfontsize}

\usepackage{pifont}
\usepackage{placeins}

\definecolor{baselineblue}{RGB}{43,124,233}
\definecolor{oralisteal}{RGB}{47,161,88}
\definecolor{qqprunorange}{RGB}{255,102,0}

\newcolumntype{Y}{>{\centering\arraybackslash}X}

\makeatletter
{\small % Capture font definitions of \small
\xdef\f@size@small{\f@size}
\xdef\f@baselineskip@small{\f@baselineskip}
\normalsize % Capture font definitions for \normalsize
\xdef\f@size@normalsize{\f@size}
\xdef\f@baselineskip@normalsize{\f@baselineskip}
}
\newcommand{\smalltonormalsize}{%
  \fontsize
    {\fpeval{(\f@size@small+\f@size@normalsize)/2}}
    {\fpeval{(\f@baselineskip@small+\f@baselineskip@normalsize)/2}}%
  \selectfont
}
\makeatother

\title{Overview of Dialog System Evaluation Track: Dimensionality, Language, Culture and Safety at DSTC 12}

\author{
  \textbf{John Mendonça\textsuperscript{1,2}},
  \textbf{Lining Zhang\textsuperscript{3}},
  \textbf{Rahul Mallidi \textsuperscript{3}},\\
  \textbf{Alon Lavie \textsuperscript{5,6}},
  \textbf{Isabel Trancoso \textsuperscript{1,2}},
  \textbf{Luis Fernando D’Haro \textsuperscript{4}},
  \textbf{João Sedoc \textsuperscript{3}}
\\
  \textsuperscript{1}INESC-ID, Lisbon \\
  \textsuperscript{2}Instituto Superior Técnico - University of Lisbon\\
  \textsuperscript{3}Department of Technology, Operations, and Statistics, New York University\\
  \textsuperscript{4}Speech Technology and Machine Learning Group - Universidad Politécnica de Madrid\\
  \textsuperscript{5}Carnegie Mellon University\\
  \textsuperscript{6}Phrase, Pittsburgh\\
  \small{
    \textbf{Correspondence:} \href{mailto:john.mendonca@inesc-id.pt}{john.mendonca@inesc-id.pt}
  }
}

\begin{document}

\maketitle

\begin{abstract}

The rapid advancement of Large Language Models (LLMs) has intensified the need for robust dialogue system evaluation, yet comprehensive assessment remains challenging. Traditional metrics often prove insufficient, and safety considerations are frequently narrowly defined or culturally biased. The DSTC12 Track 1, "Dialog System Evaluation: Dimensionality, Language, Culture and Safety," is part of the ongoing effort to address these critical gaps. The track comprised two subtasks: (1) Dialogue-level, Multi-dimensional Automatic Evaluation Metrics, and (2) Multilingual and Multicultural Safety Detection. For Task 1, focused on 10 dialogue dimensions, a Llama-3-8B baseline achieved the highest average Spearman's correlation (0.1681), indicating substantial room for improvement. In Task 2, while participating teams significantly outperformed a Llama-Guard-3-1B baseline on the multilingual safety subset (top ROC-AUC 0.9648), the baseline proved superior on the cultural subset (0.5126 ROC-AUC), highlighting critical needs in culturally-aware safety. This paper describes the datasets and baselines provided to participants, as well as submission evaluation results for each of the two proposed subtasks.

\end{abstract}

\section{Introduction}

The rapid advancements in Large Language Models (LLMs) have led to increasingly sophisticated conversational agents capable of engaging in complex and nuanced dialogues. As these models become more integrated into various applications, from customer service to personal assistants, ensuring their quality, reliability, and safety is paramount. However, evaluating dialogue systems comprehensively remains a significant challenge \citep{rodriguez-cantelar-etal-2023-overview, mendonca-etal-2024-benchmarking}. Traditional metrics often fall short of capturing the multifaceted nature of human-like conversation, and safety considerations are frequently narrowly defined or culturally biased, failing to address the full spectrum of potential issues.

Addressing the first aspect of this challenge – the limitations of current evaluation metrics – previous challenges and works focus largely on turn-level dialogue evaluation \citep{zhang2011automatic,rodriguez-cantelar-etal-2023-overview, mehri-etal-2022-interactive} and often lack further investigation of dialogue-level evaluation through automatic metrics. As LLMs advance, aspects of conversations beyond coherence, fluency, etc. should also be studied. Additionally, these aspects should provide a more fine-grained analysis of the levels of quality for the whole conversation, moving beyond simplistic turn-based scores.

\begin{table*}[tbh!]
    \centering
\begin{tabularx}{\textwidth}{|l|X|}
\hline
\textbf{Attribute} & \textbf{Description} \\
\hline
Empathy & Do you think your conversational partner had genuine empathy? \\
\hline
Trust & Based on the conversation, your conversational partner seems trustworthy \\
\hline
Skill & Based on the conversation, your conversational partner seems skilled \\
\hline
Talent & Based on the conversation, your conversational partner seems talented \\
\hline
Capability & Based on the conversation, your conversational partner seems capable \\
\hline
Relevance & Responses address the given context or query well, ensuring that the information provided is pertinent and directly applicable. \\
\hline
Non-Repetition & How repetitive was this chatbot? \\
\hline
Proactivity & Responses actively and appropriately move the conversation along different topics \\
\hline
Curiosity & How much did the chatbot try to get to know you? \\
\hline
Overall & How was the conversation? \\
\hline
\end{tabularx}
    \caption{Evaluation dimensions and definitions for Task 1.}
    \label{tab:eval-dimensions}
\end{table*}

Complementing the need for improved quality assessment, the safety dimension, highlighted as a critical concern from the outset, presents its own distinct set of urgent problems. Users are increasingly challenging current chatbots to generate harmful and/or unsafe answers. In addition, even without adversarial probing, generated responses may contain unhelpful and/or harmful content. Therefore, the automatic detection of this content is important in the deployment of these systems. Unfortunately, existing safety evaluation frameworks frequently narrow the notion of safety to strict definitions of bias and toxicity, discarding other safety aspects \citep{shuster2022blenderbot3deployedconversational,ouyang2022traininglanguagemodelsfollow}. Furthermore, a significant limitation in current safety paradigms is their predominant focus on the English language. We attempt to mitigate this bias by expanding safety datasets to a diverse set of languages and cultures. Beyond facilitating the study of safety across cultures, this also allows for the evaluation of the robustness of safety classifiers in terms of culture and language.

\subsection{Track Details}

To address these gaps, this paper presents Track 1 of DSTC12, entitled ``Dialog System Evaluation: Dimensionality, Language, Culture and Safety.'' The shared task was divided into two tasks: Dialogue-level and Multi-dimensional Automatic Evaluation Metrics (\S \ref{sec:task_1}), and Multilingual and Multicultural Safety Detection (\S \ref{sec:task_2}). This year's iteration introduced two key novelties aimed at enhancing participation and streamlining the evaluation process: (1) a focus on model efficiency and (2) the adoption of an online competition platform.

Firstly, recognizing that current dialogue evaluation research (and the broader "LLM-as-a-judge" paradigm) often relies on extremely large, proprietary models such as GPT-4 or Claude accessed via APIs, we imposed a significant constraint on model size. Participants were restricted to utilizing open-source LLMs with fewer than 13 billion parameters. This decision was motivated to encourage innovative, efficient solutions that do not solely depend on prompting state-of-the-art models. %By leveling the playing field, we aimed to shift the focus towards smarter fine-tuning strategies, novel model architectures, and effective data utilization techniques, rather than primarily prompting the largest available models.

Secondly, we utilized the Codabench platform\footnote{We have opened the competitions as benchmarks for the broader community: \href{https://www.codabench.org/competitions/7579/}{Task 1}; \href{https://www.codabench.org/competitions/7204/}{Task 2}} for managing submissions and leaderboards. This facilitated a more dynamic and interactive participation experience. We also released the datasets via Huggingface Datasets to facilitate easy access\footnote{huggingface.co/dstc12}. On the one hand, it allowed participants to easily gauge their model's performance on the development set in real-time and compare their results against established baselines. On the other hand, for the test set, participants could receive immediate feedback on their system's performance upon submission. To maintain fairness and prevent over-fitting to the test set, submissions were limited to five attempts, and the test set leaderboard remained hidden until the conclusion of the competition.

\section{Task 1: Dialogue-level and Multi-dimensional Automatic Evaluation Metrics}
\label{sec:task_1}

In this task, the goal was for participants to develop automatic evaluation metrics for open-domain dialogue. In particular, the submitted systems were expected to evaluate up to 10 different dimensions including previous common ones \cite[][i.e.]{zhang2011automatic, rodriguez-cantelar-etal-2023-overview}, together with new ones like \citep{zhang2024back}. An overview of these dimensions are presented in Table \ref{tab:eval-dimensions}. Similar to previous challenges and prior literature, we evaluated the systems using Spearman's rank correlation between human annotations and automatic metrics as our criterion.

\subsection{Dataset}

Our main dataset was separated into three collections: three bots (ChatGPT [2023], GPT-3, and BlenderBot-3) during Q1 2023 (TBD-Q1-2023), four bots (ChatGPT, Gemini, Claude, and Mixtral) during Q1 2024 (FBD-Q1-2024), and six bots (ChatGPT, Gemini, Claude, and through Hugging Chat (Mistral, Llama-3 instruct 70B, and Cohere)\footnote{The exact versions are \texttt{mistralai/Mistral-Nemo-Instruct-2407}, \texttt{meta-llama/Meta-Llama-3-70B-Instruct}, and \texttt{CohereForAI/c4ai-command-r-plus}.} during Q4 2024 (SBD-Q4-2024). The users in the conversations were undergraduate students. All conversations were read to verify that no personally identifiable information was present. For both FBD-Q1-2024 and SBD-Q4-2024 datasets, we controlled the topics present in the conversation:

\begin{itemize}
    \item [T1] Talk about help for turning your homework in late.
    \item [T2] Finding an apartment.
    \item [T3] Finding something to do in the evening.
    \item [T4] Talk about something that is on your mind or bothering you.
    \item [T5] Learn about a topic that you are interested in.
    \item [T6] Talk about something silly with the chatbot.
\end{itemize}

\noindent
Students were randomly assigned, without replacement, to both a chatbot and a conversation topic. They were instructed to interact for roughly 15 turns. After the conversation, they shared their conversation link and filled out the surveys. Subsequently, the conversation links were web scraped, and the conversational data were merged with the survey responses.

The dataset was split into development (TBD-Q1-2023 / FBD-Q1-2024) of 185 conversations and test set (SBD-Q4-2024) of 120 conversations.  TBD-Q1-2023 included 8 participants, FBD-Q1-2024 had 4, and SBD-Q4-2024 had 6.
TBD-Q1-2023 was used in the DSTC11 Track 4 challenge~\citep{rodriguez-cantelar-etal-2023-overview} for both turn- and dialog-level evaluation, but only coarse-grained dimensions were used. 

Following \citet{zhang2024back}, we used a subset of dimensions for evaluation. Table \ref{tab:eval-dimensions} has the list of dimensions along with their definitions.

\subsection{Baseline}

\begin{table*}[t]
\small
\centering
\begin{tabular}{lccccccccccc}
\toprule
\textbf{Team} &
  \textbf{Avg.} &
  \textbf{Emp.} &
  \textbf{Trust} &
  \textbf{Skill} &
  \textbf{Talent} &
  \textbf{Capab.} &
  \textbf{Rel.} &
  \textbf{!Rep.} &
  \textbf{Proact.} &
  \textbf{Curio.} &
  \textbf{Overall} \\ \midrule
Baseline & .1681 & .0647  & -.1117 & -.0955 & .0962 & .0677 & .2337 & \textit{.3851} & -.0248 & .2253 & \textit{.3766} \\
1        & .1503 & -.0788 & .0067  & -.2246 & .0529 & .1276 & .0808 & .1071 & -.1457 & \textit{.3687} & \textit{.3099} \\
2        & .1360 & -.1183 & .2407  & -.1094 & .0873 & .1717 & .0738 & .2220 & .0185  & .0064 & \textit{.3117} \\ \bottomrule
\end{tabular}
\caption{Spearman correlations for each team across all dimensions of Task 1. Correlations with $p<0.05$ in \textit{italic}.}
\label{tab:task1_res}
\end{table*}

\begin{table}[h]
\small
\centering
\begin{tabularx}{0.48\textwidth}{ X }
\vspace{0.2cm}

\cellcolor[HTML]{eeeeee}You are an impartial evaluator conducting a multidimensional assessment of text responses. Your role is to analyze and score all chatbot responses using the following criteria: \\\vspace{0.2cm}

\cellcolor[HTML]{eeeeee}- Empathy: Based on the conversation, does the chatbot demonstrate understanding and compassion for the user’s situation or emotions? \\
\cellcolor[HTML]{eeeeee}- Trust: Based on the conversation, does the chatbot seem trustworthy?\\
\cellcolor[HTML]{eeeeee}- Skill: Does the chatbot show competence in the subject matter, providing accurate and relevant information? \\
\cellcolor[HTML]{eeeeee}- Talent: Does the chatbot show talent in the subject matter, providing accurate and relevant information?\\
\cellcolor[HTML]{eeeeee}- Capability: Does the chatbot seem capable in interacting with the user?\\
\cellcolor[HTML]{eeeeee}- Relevance: Are all chatbot responses relevant given prior context?\\
\cellcolor[HTML]{eeeeee}- Non-Repetition: Does the chatbot avoid unnecessary repetition?\\
\cellcolor[HTML]{eeeeee}- Proactivity: Does the chatbot anticipate user needs?\\
\cellcolor[HTML]{eeeeee}- Curiosity: Does the chatbot demonstrate engagement by exploring the topic further or encouraging deeper discussion?  \\
\cellcolor[HTML]{eeeeee}- Non-Repetition: Does the chatbot avoid unnecessary repetition?\\
\cellcolor[HTML]{eeeeee}- Overall: Overall assessment of the chatbot throughout the dialogue.\\\vspace{0.2cm}

\cellcolor[HTML]{eeeeee}Scoring Guidelines: \\\vspace{0.001cm}

\cellcolor[HTML]{eeeeee}- Focus only on the chatbot responses, not the user messages.
\cellcolor[HTML]{eeeeee}- Assign a score between 1 and 5 for each relevant category based on the criteria above.  
\cellcolor[HTML]{eeeeee}- Do not output any other meta commentary or information.

\cellcolor[HTML]{eeeeee}Input: The input consists of a conversation between a user and a chatbot.\\\vspace{0.2cm}

\cellcolor[HTML]{eeeeee}Output: [JSON format]\\\vspace{0.2cm}

\end{tabularx}
\caption{Baseline evaluation prompt.}
\label{tab:user_evaluation}
\end{table}

As a baseline, we prompt Llama-3-8B-Instruct to provide an evaluation across all of the dimensions. The system prompt is presented in Table \ref{tab:user_evaluation}.

\subsection{Participants}

\paragraph{Team 1} Team 1 submitted four unique systems. System 1 employed a regression approach, training separate regression layers on top of a ModernBert encoder for each evaluation dimension using the DSTC-12, ConTurE \citep{ghazarian-etal-2022-wrong}, and FED \citep{mehri-eskenazi-2020-unsupervised} datasets. System 2 utilized a prompting strategy, combining detailed dimension explanations and dialogue context with a selection of models (Deepseek Llama 8B, Deepseek Qwen 7B, Qwen 2.5 7B Instruct-1M), choosing the best-performing model per dimension based on validation set results. System 3 was a classification-based approach, training individual classifiers on an sBERT encoder for each dimension with normalized scores, also using the DSTC-12, ConTurE, and FED datasets. Finally, System 4, a hybrid model, selectively combined the outputs of System 1 (for dimensions like Talent and Relevance) and System 2 (for dimensions like Empathy and Overall) based on which system achieved the best correlation on the validation set for each specific dimension.

\paragraph{Team 2.} This team adopted Qwen2.5-7B-Instruct as the base model and then utilized prompt engineering to enable the LLM to automatically output scores across various dimensions. Moreover, they included degree interpretations for different score levels within the prompt.

\subsection{Results}

The official results for Task 1 are provided in Table \ref{tab:task1_res}. The team score was computed as the mean absolute Spearman correlation across all dimensions. We can also see a per-dimension breakdown. Ideally, all correlations should be positive; however, in the absence of consistent definitions, some participants may have reverse-coded certain dimensions.

Quite surprisingly, the baseline outperformed the other teams, followed by Team 1 and then Team 2. This was largely due to the dimensions of relevance and non-repetition. While the baseline performed best overall, there was no clear winner across all dimensions, indicating substantial room for improvement.  Most correlations were statistically significant, though relatively low for several dimensions. Unsurprisingly, all metrics correlated well with the overall conversational quality. The negative correlation for the 'skill' dimension highlights a clear area for further refinement.

We had one strange finding: on a relative basis, all automatic metrics favored Mistral, whereas human participants ranked it significantly lower. This is probably the reason for the large decrease in correlations between development and test sets.

\section{Task 2: Multilingual and Multicultural Safety Detection}
\label{sec:task_2}

In this task, participants were tasked with developing automatic safety classifiers of responses generated by LLMs across different languages and cultures. To establish a clear and manageable benchmark, particularly given the diverse range of languages and the nuanced nature of safety, we simplified the core challenge to a binary classification problem: determining whether a given dialogue response is 'safe' or 'unsafe'. We consider a response to be "unsafe" if it violates the MLCommons AI Safety v0.5 Benchmark POC Taxonomy of Hazards \citep{vidgen2024introducingv05aisafety}\footnote{\url{https://drive.google.com/file/d/1V8KFfk8awaAXc83nZZzDV2bHgPT8jbJY/view}}. The taxonomy outlines seven key hazard categories within its scope for this version: Violent crimes, Non-violent crimes, Sex-related crimes, Child sexual exploitation, Indiscriminate weapons (CBRNE), Suicide \& self-harm, and Hate. This approach, while a simplification of real-world complexities where harm can be multi-faceted and context-dependent, allows for a more direct comparison of model capabilities in fundamental safety detection across varied linguistic and cultural contexts. It also provides a foundational step before tackling more granular multi-label or severity-level classifications.

\begin{table}[t]
\centering
\begin{tabular}{lc}
\toprule
Dataset            & \#Utterances (k)    \\ \midrule
BAD                & 69.3 / 7 / 2.6      \\
Dialogue Safety    & 24 / 3 / 3          \\
Prosocial Dialogue & 120 / 20.4 / 25     \\
Total              & 213.3 / 30.4 / 30.6 \\ \midrule \midrule
SODA-Eval          & - / - / 325         \\
CoSafe             & - / - / 227         \\
SafeWorld          & - / - / 437         \\ \bottomrule
\end{tabular}
\caption{Overview of datasets used in Task 2. For the development set, we provide train/validation/test sets.}
\label{tab:data_task2}
\end{table}

\subsection{Datasets}

To allow participants to train and evaluate their models, we curated several datasets. These datasets were processed to conform to a shared conversational format, consisting of context and response pairs accompanied with additional metadata made available in the original datasets. These datasets were then translated to 7 languages (Arabic, German, Spanish, French, Japanese, Portuguese and Chinese) and made accessible to the participants on HuggingFace\footnote{\url{https://huggingface.co/dstc12}}. We present an overview of these datasets in Table \ref{tab:data_task2}.

\subsubsection{Development}

\paragraph{Bot Adversarial Dialogue \citep{xu-etal-2021-bot}.}

This dataset was curated via a human-and-model-in-the-loop framework where crowdworkers were instructed to converse with various state-of-the-art dialogue models, actively probing the model to output unsafe or offensive responses. Each bot utterance within these interactions was annotated for safety, resulting in a corpus of approximately 5.8k dialogues (79k total utterances), with 40\% of utterances being annotated as offensive.

\paragraph{Dialogue Safety \citep{dinan-etal-2019-build}} was curated via a human-and-model-in-the-loop framework. Crowdworkers were presented with an existing dialogue context and were instructed to submit utterances they deemed offensive, specifically targeting those that an existing safety classifier would miss-classify as safe. This iterative process resulted in a corpus of approximately 6,000 "offensive" utterances, collected across both single-turn and multi-turn dialogue context settings. When combined with verified safe examples, these constitute a dataset totalling approximately 60,000 utterances, of which 10\% are labelled offensive. For the purpose of this Task, we use the multi-turn subset.

\paragraph{ProsocialDialog \citep{kim-etal-2022-prosocialdialog}} 

is a large-scale, multi-turn English dialogue dataset designed to teach conversational agents to respond prosocially to problematic user inputs. Generated via a human-AI collaborative framework, it contains 58,137 dialogues (331,362 utterances) covering diverse unethical, problematic, biased, and toxic situations. Prosocial responses are grounded in 160,295 commonsense social rules-of-thumb (RoTs), and dialogue turns are annotated with fine-grained safety labels accompanied by 497,043 free-form rationales.

\subsubsection{Test}

\paragraph{Soda-Eval \citep{mendonca-etal-2024-soda}} is derived from the SODA dataset, and encompasses over 120,000 turn-level assessments across 10,000 dialogues. Each assessment, generated by GPT-4 and subsequently human-validated, includes identification of fine-grained issues. We leverage the annotations for the quality dimensions "Anti-Social". However, since these annotations were automatically annotated using an LLM (GPT-4), we conduct a human validation step on the safety labels. From this validation step, we randomly select additional positive examples from Soda-Eval to derive a class-balanced set of size 325 examples.

\paragraph{CoSafe \citep{yu-etal-2024-cosafe}} is a benchmark designed to evaluate safety against multi-turn dialogue coreference attacks. The dataset was constructed by selecting 100 single-turn attack prompts for each of 14 harmful categories, originally defined by BeaverTails (Ji et al., 2023). These prompts were then automatically expanded into multi-turn dialogues using GPT-4, with the coreferentially-phrased attack query placed in the final turn to assess model vulnerabilities in resolving references within a harmful conversational context. We employed multiple LLMs to simulate diverse safety behaviors across model families and architectures. This diversity ensures that safety classifiers are not overfitted to idiosyncrasies of a single generation style and that evaluation generalizes across real-world deployment scenarios. The chosen models are aya-expanse-8b \citep{dang2024ayaexpansecombiningresearch}, EuroLLM-9B-instruct  \citep{martins2024eurollmmultilinguallanguagemodels}, LLama-3.2-Instruct (1B,3B) and LLama-3.1-8B-Instruct \citep{grattafiori2024llama3herdmodels}, Ministral-8B-Instruct-2410 \citep{ministral}, and Qwen2.5-Instruct (3B,7B) \citep{qwen2025qwen25technicalreport}. We then conduct a human-validated automated annotation using GPT-4o \citep{gpt4omini} as an automated safety classifier. Then, all examples rated as unsafe are evaluated by a human annotator. A balanced safety-label subset is then sampled from these annotations.

\paragraph{SafeWorld \citep{yin2024safeworld}} For the cultural subtask, we employ a curated version of the cultural-aware safety dataset of SafeWorld \citep{yin2024safeworld}. SafeWorld is designed to assess alignment with geo-diverse cultural and legal safety standards by grounding queries on human-verified cultural norms and legal policies from 50 countries and 493 distinct regions/races. We focus on the "specific answer" and "comprehensive answer" query types. "Specific answer" queries (641 instances) require models to pinpoint a single, pre-defined cultural or legal guideline violated in a given scenario; "comprehensive answer" queries (577 instances) present situations where potential violations are ambiguous, tasking models to provide comprehensive responses covering relevant norms and policies across implicated regions. We prompt GPT-4o to determine if the policy or norm violation identified would elicit a safety violation. For the examples identified as unsafe, we then generate responses using several LLMs: aya-expanse-8B \citep{dang2024ayaexpansecombiningresearch}, EuroLLM-9B-instruct \citep{martins2024eurollmmultilinguallanguagemodels}, gemma-3-9b-it \citep{gemmateam2025gemma3technicalreport}; gpt-4o-mini \citep{gpt4omini}, LLama-3.1-8B-Instruct \citep{grattafiori2024llama3herdmodels}, and Qwen2.57B-Instruct \citep{qwen2025qwen25technicalreport}. Then, we ask GPT-4o \citep{gpt4omini} to determine if the the model response elicits the identified policy/norm violation.

Human validations to confirm the accuracy of the test set labels was conducted by a single annotator, a graduate student with a background of Computational Linguistics with extensive annotation experience.

\subsubsection{Translations}

We translate all of the context-response pairs for the development set as a whole, and for Soda-Eval and CoSafe. We selected a wide-range of languages that cover a diverse set of cultures: Arabic, German, English, Spanish, French, Japanese, Portuguese and Chinese.

\paragraph{Automated validation} For translation model selection, we automatically validate the quality of translations in a smaller subset by employing an MT-QE model, in this case \texttt{wmt23-cometkiwi-da-xl}. We obtain a score of 0.7116 using GPT-4o and 0.7153 GPT-4o-mini. Given the small performance difference and the much lower price point for GPT-4o-mini, we opt with utilizing GPT-4o-mini for the translation of the full development set.

\paragraph{Human validation} For the test set, we additionally conduct a human validation study to ensure the translations are of high quality and do not change the semantics of the response such that it changes the safety label. Human annotators -- one per language, except for Japanese (2) -- were recruited given their professional level of the target language and consisted of a mix of graduate students and post-docs with Linguistics and Computational Linguistics background with prior experience with annotations. Initial annotations showed that GPT-4o-mini was incurring in several translation errors that affect the safety of the response. As such, we retranslate the English set with GPT-4o, which, after subsequent human validation, confirmed the quality of its translations on the examples GPT-4o-mini failed.

\subsection{Baseline}

We adopted Llama-Guard-3-1B as the official baseline for Task 2. This model was aligned from the original Llama-3.2-1B LLM to safeguard against the MLCommons standardized hazards taxonomy. In particular, the English data used by Llama Guard was combined with additional multilingual human and synthetically generated Human-AI conversation data.

For the purposes of the task, baseline predictions for response classification are obtained by extracting the probability of the "unsafe" token appearing as the first decoded token to reflect the model’s direct scoring of risk. This approach offers a deterministic, resource-efficient proxy for classification, aligning with recent work in zero-shot risk estimation and avoiding confounding artifacts from longer generation chains. Reproducible code can be found alongside the data on the HuggingFace dataset repository\footnote{\url{https://huggingface.co/datasets/dstc12/bot_adversarial_dialogue/blob/main/LlamaGuard.py}}.

\subsection{Participants}

For this task, a total of 2 teams (Teams 2 and 3) participated with 6 submissions. Participants were asked to provide a description of their submissions. Team 2 submitted a similar system to the one presented in Task 1 (\ref{sec:task_1}, adapting the prompt for the safety task. Unfortunately, Team 3 did not provide an official description of their system. However, their submissions to the track platform suggest their approach consisted in the supervised finetuning of LLMs on the development data (\texttt{sft\_500k\_gemma\-ck} and \texttt{llama3\_sft\_500k}) of gemma-2-9b-it and a LLama3 model respecting our model size restrictions (likely 8B).

\begin{table}[h]
\centering
\
\begin{tabular}{lccc}
\toprule
\textbf{Team}     & \textbf{Average} & \textbf{Cultural} & \textbf{Multilingual} \\ \midrule
3 & \textbf{.9046}       & \underline{.4831}     & \textbf{.9648}        \\
2 & \underline{.8078}    & \textit{.4830}        & \underline{.8517}        \\
Baseline & \textit{.7767}       & \textbf{.5126}        & \textit{.8097}      \\\bottomrule
\end{tabular}
\caption{ROC-AUC results for Task 2. The first position is shown in \textbf{bold}, the second in \underline{underline} and the third in \textit{italic}.}
\label{tab:task2}
\end{table}

\subsection{Results}

The official results for Task 2 are provided in Table \ref{tab:task2}. Team ranking is established by calculating the average ROC-AUC considering all languages and the cultural subset with equal weights. We also present ROC-AUC for the multilingual and cultural subsets separately. We employ ROC-AUC since it provides a threshold-independent assessment of a model's ability to distinguish between safe and unsafe content.

Team 3 ranked best in this Task, followed by Team 2. This is thanks to their strong performance on the multilingual subset, with Team 3 achieving a strong result of .9648, followed by Team 2 with .8517, which are significantly superior to the baseline results (.8097). However, when looking at the cultural subset, we note that the baseline was the best performing submission (.5126), with both Teams achieving similar results (around .4831). These results suggest that models finetuned for cultural agnostic safety concerns fail to account for cultural specificities. This behaviour may be an instance of catastrophic forgetting, since our baseline (LLama-Guard-3-1B) was able to outperform their stronger finetuned models.

\section{Conclusions and Future Work}

This paper presents the overview of Track 1 on "Dialog System Evaluation: Dimensionality, Language, Culture and Safety" organized as part of the 12th Dialogue System Technology Challenge (DSTC12). The track was organized in two tasks aimed at addressing two important problems of the state-of-the-art in Dialogue Systems: (1) Dialogue-level and Multi-dimensional Automatic Evaluation Metrics; (2) Multilingual and Multicultural Safety Detection.

While the track had 11 registered teams, only 3 participated. The first task drew two of these teams. We used Spearman's rank correlation coefficient absolute average value as the rank ordering for the teams. The baseline outperformed the best overall, but alone many different dimensions we see different methods performing better. 

In the second task, two teams participated and comfortably outperformed the baseline on the multilingual subset, achieving very strong ROC-AUC. However, for the cultural subset, no team was able to outperform the baseline ROC-AUC, which sits at just .5126, indicating clear room for improvement.

As future work, Task 1, we plan to extend the analysis of fine-grained dimensions to understand the upper-bound of LLM-evaluation for dimensions of human quality assessment. Importantly, we plan to increase the diversity of participants to be more representative of larger populations. For Task 2, we plan on extending the safety classification task to include the full taxonomy, providing a more fine-grained assessment of risks.

\section*{Acknowledgments}

This research was supported by the Portuguese Recovery and Resilience Plan through project C645008882-00000055 (Responsible.AI), by Portuguese national funds through Fundação para a Ciência e Tecnologia (FCT) with references PRT/BD/152198/2021 and DOI:10.54499/UIDB/50021/2020.

This work is supported by the European Commission through Project ASTOUND (101071191 -- HORIZON EIC-2021 -- PATHFINDERCHALLENGES-01), and by project BEWORD (PID2021-126061OB-C43) funded by MCIN/AEI/10.13039/501100011033 and, as appropriate, by "ERDF A way of making Europe", by the European Union.

We also want to give thanks to MS Azure services (especially to Irving Kwong) for their sponsorship to continue processing new datasets that could be interesting for the dialogue community.

This research project is supported by the NYU ChatEval Team led by João Sedoc. He would like to thank NYU Stern for its funding.

\bibliography{custom,anthology_0,anthology_1}

\begin{thebibliography}{23}
\providecommand{\natexlab}[1]{#1}

\bibitem[{Dang et~al.(2024)Dang, Singh, D'souza, Ahmadian, Salamanca, Smith, Peppin, Hong, Govindassamy, Zhao, Kublik, Amer, Aryabumi, Campos, Tan, Kocmi, Strub, Grinsztajn, Flet-Berliac, Locatelli, Lin, Talupuru, Venkitesh, Cairuz, Yang, Chung, Ko, Shi, Shukayev, Bae, Piktus, Castagné, Cruz-Salinas, Kim, Crawhall-Stein, Morisot, Roy, Blunsom, Zhang, Gomez, Frosst, Fadaee, Ermis, Üstün, and Hooker}]{dang2024ayaexpansecombiningresearch}
John Dang, Shivalika Singh, Daniel D'souza, Arash Ahmadian, Alejandro Salamanca, Madeline Smith, Aidan Peppin, Sungjin Hong, Manoj Govindassamy, Terrence Zhao, Sandra Kublik, Meor Amer, Viraat Aryabumi, Jon~Ander Campos, Yi-Chern Tan, Tom Kocmi, Florian Strub, Nathan Grinsztajn, Yannis Flet-Berliac, and 26 others. 2024.
\newblock \href {https://arxiv.org/abs/2412.04261} {{Aya Expanse: Combining Research Breakthroughs for a New Multilingual Frontier}}.
\newblock \emph{Preprint}, arXiv:2412.04261.

\bibitem[{Dinan et~al.(2019)Dinan, Humeau, Chintagunta, and Weston}]{dinan-etal-2019-build}
Emily Dinan, Samuel Humeau, Bharath Chintagunta, and Jason Weston. 2019.
\newblock \href {https://doi.org/10.18653/v1/D19-1461} {Build it break it fix it for dialogue safety: Robustness from adversarial human attack}.
\newblock In \emph{Proceedings of the 2019 Conference on Empirical Methods in Natural Language Processing and the 9th International Joint Conference on Natural Language Processing (EMNLP-IJCNLP)}, pages 4537--4546, Hong Kong, China. Association for Computational Linguistics.

\bibitem[{Gemma et~al.(2025)Gemma, Kamath, Ferret, Pathak, Vieillard, Merhej, Perrin, Matejovicova, Ramé, Rivière, Rouillard, Mesnard, Cideron, bastien Grill, Ramos, Yvinec, Casbon, Pot, Penchev, Liu, Visin, Kenealy, Beyer, Zhai, Tsitsulin, Busa-Fekete, Feng, Sachdeva, Coleman, Gao, Mustafa, Barr, Parisotto, Tian, Eyal, Cherry, Peter, Sinopalnikov, Bhupatiraju, Agarwal, Kazemi, Malkin, Kumar, Vilar, Brusilovsky, Luo, Steiner, Friesen, Sharma, Sharma, Gilady, Goedeckemeyer, Saade, Feng, Kolesnikov, Bendebury, Abdagic, Vadi, György, Pinto, Das, Bapna, Miech, Yang, Paterson, Shenoy, Chakrabarti, Piot, Wu, Shahriari, Petrini, Chen, Lan, Choquette-Choo, Carey, Brick, Deutsch, Eisenbud, Cattle, Cheng, Paparas, Sreepathihalli, Reid, Tran, Zelle, Noland, Huizenga, Kharitonov, Liu, Amirkhanyan, Cameron, Hashemi, Klimczak-Plucińska, Singh, Mehta, Lehri, Hazimeh, Ballantyne, Szpektor, Nardini, Pouget-Abadie, Chan, Stanton, Wieting, Lai, Orbay, Fernandez, Newlan, yeong Ji, Singh, Black, Yu, Hui, Vodrahalli, Greff,
  Qiu, Valentine, Coelho, Ritter, Hoffman, Watson, Chaturvedi, Moynihan, Ma, Babar, Noy, Byrd, Roy, Momchev, Chauhan, Sachdeva, Bunyan, Botarda, Caron, Rubenstein, Culliton, Schmid, Sessa, Xu, Stanczyk, Tafti, Shivanna, Wu, Pan, Rokni, Willoughby, Vallu, Mullins, Jerome, Smoot, Girgin, Iqbal, Reddy, Sheth, Põder, Bhatnagar, Panyam, Eiger, Zhang, Liu, Yacovone, Liechty, Kalra, Evci, Misra, Roseberry, Feinberg, Kolesnikov, Han, Kwon, Chen, Chow, Zhu, Wei, Egyed, Cotruta, Giang, Kirk, Rao, Black, Babar, Lo, Moreira, Martins, Sanseviero, Gonzalez, Gleicher, Warkentin, Mirrokni, Senter, Collins, Barral, Ghahramani, Hadsell, Matias, Sculley, Petrov, Fiedel, Shazeer, Vinyals, Dean, Hassabis, Kavukcuoglu, Farabet, Buchatskaya, Alayrac, Anil, Dmitry, Lepikhin, Borgeaud, Bachem, Joulin, Andreev, Hardin, Dadashi, and Hussenot}]{gemmateam2025gemma3technicalreport}
Team Gemma, Aishwarya Kamath, Johan Ferret, Shreya Pathak, Nino Vieillard, Ramona Merhej, Sarah Perrin, Tatiana Matejovicova, Alexandre Ramé, Morgane Rivière, Louis Rouillard, Thomas Mesnard, Geoffrey Cideron, Jean bastien Grill, Sabela Ramos, Edouard Yvinec, Michelle Casbon, Etienne Pot, Ivo Penchev, and 197 others. 2025.
\newblock \href {https://arxiv.org/abs/2503.19786} {{Gemma 3 Technical Report}}.
\newblock \emph{Preprint}, arXiv:2503.19786.

\bibitem[{Ghazarian et~al.(2022)Ghazarian, Hedayatnia, Papangelis, Liu, and Hakkani-Tur}]{ghazarian-etal-2022-wrong}
Sarik Ghazarian, Behnam Hedayatnia, Alexandros Papangelis, Yang Liu, and Dilek Hakkani-Tur. 2022.
\newblock \href {https://doi.org/10.18653/v1/2022.findings-acl.331} {What is wrong with you?: Leveraging user sentiment for automatic dialog evaluation}.
\newblock In \emph{Findings of the Association for Computational Linguistics: ACL 2022}, pages 4194--4204, Dublin, Ireland. Association for Computational Linguistics.

\bibitem[{Grattafiori et~al.(2024)Grattafiori, Dubey, Jauhri, Pandey, Kadian, Al-Dahle, Letman, Mathur, Schelten, Vaughan, Yang, Fan, Goyal, Hartshorn, Yang, Mitra, Sravankumar, Korenev, Hinsvark, Rao, Zhang, Rodriguez, Gregerson, Spataru, Roziere, Biron, Tang, Chern, Caucheteux, Nayak, Bi, Marra, McConnell, Keller, Touret, Wu, Wong, Ferrer, Nikolaidis, Allonsius, Song, Pintz, Livshits, Wyatt, Esiobu, Choudhary, Mahajan, Garcia-Olano, Perino, Hupkes, Lakomkin, AlBadawy, Lobanova, Dinan, Smith, Radenovic, Guzmán, Zhang, Synnaeve, Lee, Anderson, Thattai, Nail, Mialon, Pang, Cucurell, Nguyen, Korevaar, Xu, Touvron, Zarov, Ibarra, Kloumann, Misra, Evtimov, Zhang, Copet, Lee, Geffert, Vranes, Park, Mahadeokar, Shah, van~der Linde, Billock, Hong, Lee, Fu, Chi, Huang, Liu, Wang, Yu, Bitton, Spisak, Park, Rocca, Johnstun, Saxe, Jia, Alwala, Prasad, Upasani, Plawiak, Li, Heafield, Stone, El-Arini, Iyer, Malik, Chiu, Bhalla, Lakhotia, Rantala-Yeary, van~der Maaten, Chen, Tan, Jenkins, Martin, Madaan, Malo, Blecher,
  Landzaat, de~Oliveira, Muzzi, Pasupuleti, Singh, Paluri, Kardas, Tsimpoukelli, Oldham, Rita, Pavlova, Kambadur, Lewis, Si, Singh, Hassan, Goyal, Torabi, Bashlykov, Bogoychev, Chatterji, Zhang, Duchenne, Çelebi, Alrassy, Zhang, Li, Vasic, Weng, Bhargava, Dubal, Krishnan, Koura, Xu, He, Dong, Srinivasan, Ganapathy, Calderer, Cabral, Stojnic, Raileanu, Maheswari, Girdhar, Patel, Sauvestre, Polidoro, Sumbaly, Taylor, Silva, Hou, Wang, Hosseini, Chennabasappa, Singh, Bell, Kim, Edunov, Nie, Narang, Raparthy, Shen, Wan, Bhosale, Zhang, Vandenhende, Batra, Whitman, Sootla, Collot, Gururangan, Borodinsky, Herman, Fowler, Sheasha, Georgiou, Scialom, Speckbacher, Mihaylov, Xiao, Karn, Goswami, Gupta, Ramanathan, Kerkez, Gonguet, Do, Vogeti, Albiero, Petrovic, Chu, Xiong, Fu, Meers, Martinet, Wang, Wang, Tan, Xia, Xie, Jia, Wang, Goldschlag, Gaur, Babaei, Wen, Song, Zhang, Li, Mao, Coudert, Yan, Chen, Papakipos, Singh, Srivastava, Jain, Kelsey, Shajnfeld, Gangidi, Victoria, Goldstand, Menon, Sharma, Boesenberg,
  Baevski, Feinstein, Kallet, Sangani, Teo, Yunus, Lupu, Alvarado, Caples, Gu, Ho, Poulton, Ryan, Ramchandani, Dong, Franco, Goyal, Saraf, Chowdhury, Gabriel, Bharambe, Eisenman, Yazdan, James, Maurer, Leonhardi, Huang, Loyd, Paola, Paranjape, Liu, Wu, Ni, Hancock, Wasti, Spence, Stojkovic, Gamido, Montalvo, Parker, Burton, Mejia, Liu, Wang, Kim, Zhou, Hu, Chu, Cai, Tindal, Feichtenhofer, Gao, Civin, Beaty, Kreymer, Li, Adkins, Xu, Testuggine, David, Parikh, Liskovich, Foss, Wang, Le, Holland, Dowling, Jamil, Montgomery, Presani, Hahn, Wood, Le, Brinkman, Arcaute, Dunbar, Smothers, Sun, Kreuk, Tian, Kokkinos, Ozgenel, Caggioni, Kanayet, Seide, Florez, Schwarz, Badeer, Swee, Halpern, Herman, Sizov, Guangyi, Zhang, Lakshminarayanan, Inan, Shojanazeri, Zou, Wang, Zha, Habeeb, Rudolph, Suk, Aspegren, Goldman, Zhan, Damlaj, Molybog, Tufanov, Leontiadis, Veliche, Gat, Weissman, Geboski, Kohli, Lam, Asher, Gaya, Marcus, Tang, Chan, Zhen, Reizenstein, Teboul, Zhong, Jin, Yang, Cummings, Carvill, Shepard, McPhie,
  Torres, Ginsburg, Wang, Wu, U, Saxena, Khandelwal, Zand, Matosich, Veeraraghavan, Michelena, Li, Jagadeesh, Huang, Chawla, Huang, Chen, Garg, A, Silva, Bell, Zhang, Guo, Yu, Moshkovich, Wehrstedt, Khabsa, Avalani, Bhatt, Mankus, Hasson, Lennie, Reso, Groshev, Naumov, Lathi, Keneally, Liu, Seltzer, Valko, Restrepo, Patel, Vyatskov, Samvelyan, Clark, Macey, Wang, Hermoso, Metanat, Rastegari, Bansal, Santhanam, Parks, White, Bawa, Singhal, Egebo, Usunier, Mehta, Laptev, Dong, Cheng, Chernoguz, Hart, Salpekar, Kalinli, Kent, Parekh, Saab, Balaji, Rittner, Bontrager, Roux, Dollar, Zvyagina, Ratanchandani, Yuvraj, Liang, Alao, Rodriguez, Ayub, Murthy, Nayani, Mitra, Parthasarathy, Li, Hogan, Battey, Wang, Howes, Rinott, Mehta, Siby, Bondu, Datta, Chugh, Hunt, Dhillon, Sidorov, Pan, Mahajan, Verma, Yamamoto, Ramaswamy, Lindsay, Lindsay, Feng, Lin, Zha, Patil, Shankar, Zhang, Zhang, Wang, Agarwal, Sajuyigbe, Chintala, Max, Chen, Kehoe, Satterfield, Govindaprasad, Gupta, Deng, Cho, Virk, Subramanian, Choudhury,
  Goldman, Remez, Glaser, Best, Koehler, Robinson, Li, Zhang, Matthews, Chou, Shaked, Vontimitta, Ajayi, Montanez, Mohan, Kumar, Mangla, Ionescu, Poenaru, Mihailescu, Ivanov, Li, Wang, Jiang, Bouaziz, Constable, Tang, Wu, Wang, Wu, Gao, Kleinman, Chen, Hu, Jia, Qi, Li, Zhang, Zhang, Adi, Nam, Yu, Wang, Zhao, Hao, Qian, Li, He, Rait, DeVito, Rosnbrick, Wen, Yang, Zhao, and Ma}]{grattafiori2024llama3herdmodels}
Aaron Grattafiori, Abhimanyu Dubey, Abhinav Jauhri, Abhinav Pandey, Abhishek Kadian, Ahmad Al-Dahle, Aiesha Letman, Akhil Mathur, Alan Schelten, Alex Vaughan, Amy Yang, Angela Fan, Anirudh Goyal, Anthony Hartshorn, Aobo Yang, Archi Mitra, Archie Sravankumar, Artem Korenev, Arthur Hinsvark, and 542 others. 2024.
\newblock \href {https://arxiv.org/abs/2407.21783} {{The Llama 3 Herd of Models}}.
\newblock \emph{Preprint}, arXiv:2407.21783.

\bibitem[{Kim et~al.(2022)Kim, Yu, Jiang, Lu, Khashabi, Kim, Choi, and Sap}]{kim-etal-2022-prosocialdialog}
Hyunwoo Kim, Youngjae Yu, Liwei Jiang, Ximing Lu, Daniel Khashabi, Gunhee Kim, Yejin Choi, and Maarten Sap. 2022.
\newblock \href {https://doi.org/10.18653/v1/2022.emnlp-main.267} {{P}rosocial{D}ialog: A prosocial backbone for conversational agents}.
\newblock In \emph{Proceedings of the 2022 Conference on Empirical Methods in Natural Language Processing}, pages 4005--4029, Abu Dhabi, United Arab Emirates. Association for Computational Linguistics.

\bibitem[{Martins et~al.(2024)Martins, Fernandes, Alves, Guerreiro, Rei, Alves, Pombal, Farajian, Faysse, Klimaszewski, Colombo, Haddow, de~Souza, Birch, and Martins}]{martins2024eurollmmultilinguallanguagemodels}
Pedro~Henrique Martins, Patrick Fernandes, João Alves, Nuno~M. Guerreiro, Ricardo Rei, Duarte~M. Alves, José Pombal, Amin Farajian, Manuel Faysse, Mateusz Klimaszewski, Pierre Colombo, Barry Haddow, José G.~C. de~Souza, Alexandra Birch, and André F.~T. Martins. 2024.
\newblock \href {https://arxiv.org/abs/2409.16235} {Eurollm: Multilingual language models for europe}.
\newblock \emph{Preprint}, arXiv:2409.16235.

\bibitem[{Mehri and Eskenazi(2020)}]{mehri-eskenazi-2020-unsupervised}
Shikib Mehri and Maxine Eskenazi. 2020.
\newblock \href {https://doi.org/10.18653/v1/2020.sigdial-1.28} {Unsupervised evaluation of interactive dialog with {D}ialo{GPT}}.
\newblock In \emph{Proceedings of the 21th Annual Meeting of the Special Interest Group on Discourse and Dialogue}, pages 225--235, 1st virtual meeting. Association for Computational Linguistics.

\bibitem[{Mehri et~al.(2022)Mehri, Feng, Gordon, Alavi, Traum, and Eskenazi}]{mehri-etal-2022-interactive}
Shikib Mehri, Yulan Feng, Carla Gordon, Seyed~Hossein Alavi, David Traum, and Maxine Eskenazi. 2022.
\newblock \href {https://aclanthology.org/2022.lrec-1.616/} {Interactive evaluation of dialog track at {DSTC}9}.
\newblock In \emph{Proceedings of the Thirteenth Language Resources and Evaluation Conference}, pages 5731--5738, Marseille, France. European Language Resources Association.

\bibitem[{Mendon{\c{c}}a et~al.(2024{\natexlab{a}})Mendon{\c{c}}a, Lavie, and Trancoso}]{mendonca-etal-2024-benchmarking}
John Mendon{\c{c}}a, Alon Lavie, and Isabel Trancoso. 2024{\natexlab{a}}.
\newblock \href {https://aclanthology.org/2024.nlp4convai-1.1/} {On the benchmarking of {LLM}s for open-domain dialogue evaluation}.
\newblock In \emph{Proceedings of the 6th Workshop on NLP for Conversational AI (NLP4ConvAI 2024)}, pages 1--12, Bangkok, Thailand. Association for Computational Linguistics.

\bibitem[{Mendon{\c{c}}a et~al.(2024{\natexlab{b}})Mendon{\c{c}}a, Trancoso, and Lavie}]{mendonca-etal-2024-soda}
John Mendon{\c{c}}a, Isabel Trancoso, and Alon Lavie. 2024{\natexlab{b}}.
\newblock \href {https://doi.org/10.18653/v1/2024.findings-emnlp.684} {Soda-eval: Open-domain dialogue evaluation in the age of {LLM}s}.
\newblock In \emph{Findings of the Association for Computational Linguistics: EMNLP 2024}, pages 11687--11708, Miami, Florida, USA. Association for Computational Linguistics.

\bibitem[{MistralAI(2024)}]{ministral}
MistralAI. 2024.
\newblock \href {https://mistral.ai/news/ministraux} {Un ministral, des ministraux | mistral ai}.

\bibitem[{OpenAI et~al.(2024)OpenAI, Hurst, Lerer, Goucher, Perelman, Ramesh, Clark, Ostrow, Welihinda, Hayes, Radford, Mądry, Baker-Whitcomb, Beutel, Borzunov, Carney, Chow, Kirillov, Nichol, Paino, Renzin, Passos, Kirillov, Christakis, Conneau, Kamali, Jabri, Moyer, Tam, Crookes, Tootoochian, Tootoonchian, Kumar, Vallone, Karpathy, Braunstein, Cann, Codispoti, Galu, Kondrich, Tulloch, Mishchenko, Baek, Jiang, Pelisse, Woodford, Gosalia, Dhar, Pantuliano, Nayak, Oliver, Zoph, Ghorbani, Leimberger, Rossen, Sokolowsky, Wang, Zweig, Hoover, Samic, McGrew, Spero, Giertler, Cheng, Lightcap, Walkin, Quinn, Guarraci, Hsu, Kellogg, Eastman, Lugaresi, Wainwright, Bassin, Hudson, Chu, Nelson, Li, Shern, Conger, Barette, Voss, Ding, Lu, Zhang, Beaumont, Hallacy, Koch, Gibson, Kim, Choi, McLeavey, Hesse, Fischer, Winter, Czarnecki, Jarvis, Wei, Koumouzelis, Sherburn, Kappler, Levin, Levy, Carr, Farhi, Mely, Robinson, Sasaki, Jin, Valladares, Tsipras, Li, Nguyen, Findlay, Oiwoh, Wong, Asdar, Proehl, Yang, Antonow,
  Kramer, Peterson, Sigler, Wallace, Brevdo, Mays, Khorasani, Such, Raso, Zhang, von Lohmann, Sulit, Goh, Oden, Salmon, Starace, Brockman, Salman, Bao, Hu, Wong, Wang, Schmidt, Whitney, Jun, Kirchner, de~Oliveira~Pinto, Ren, Chang, Chung, Kivlichan, O'Connell, O'Connell, Osband, Silber, Sohl, Okuyucu, Lan, Kostrikov, Sutskever, Kanitscheider, Gulrajani, Coxon, Menick, Pachocki, Aung, Betker, Crooks, Lennon, Kiros, Leike, Park, Kwon, Phang, Teplitz, Wei, Wolfe, Chen, Harris, Varavva, Lee, Shieh, Lin, Yu, Weng, Tang, Yu, Jang, Candela, Beutler, Landers, Parish, Heidecke, Schulman, Lachman, McKay, Uesato, Ward, Kim, Huizinga, Sitkin, Kraaijeveld, Gross, Kaplan, Snyder, Achiam, Jiao, Lee, Zhuang, Harriman, Fricke, Hayashi, Singhal, Shi, Karthik, Wood, Rimbach, Hsu, Nguyen, Gu-Lemberg, Button, Liu, Howe, Muthukumar, Luther, Ahmad, Kai, Itow, Workman, Pathak, Chen, Jing, Guy, Fedus, Zhou, Mamitsuka, Weng, McCallum, Held, Ouyang, Feuvrier, Zhang, Kondraciuk, Kaiser, Hewitt, Metz, Doshi, Aflak, Simens, Boyd,
  Thompson, Dukhan, Chen, Gray, Hudnall, Zhang, Aljubeh, Litwin, Zeng, Johnson, Shetty, Gupta, Shah, Yatbaz, Yang, Zhong, Glaese, Chen, Janner, Lampe, Petrov, Wu, Wang, Fradin, Pokrass, Castro, de~Castro, Pavlov, Brundage, Wang, Khan, Murati, Bavarian, Lin, Yesildal, Soto, Gimelshein, Cone, Staudacher, Summers, LaFontaine, Chowdhury, Ryder, Stathas, Turley, Tezak, Felix, Kudige, Keskar, Deutsch, Bundick, Puckett, Nachum, Okelola, Boiko, Murk, Jaffe, Watkins, Godement, Campbell-Moore, Chao, McMillan, Belov, Su, Bak, Bakkum, Deng, Dolan, Hoeschele, Welinder, Tillet, Pronin, Tillet, Dhariwal, Yuan, Dias, Lim, Arora, Troll, Lin, Lopes, Puri, Miyara, Leike, Gaubert, Zamani, Wang, Donnelly, Honsby, Smith, Sahai, Ramchandani, Huet, Carmichael, Zellers, Chen, Chen, Nigmatullin, Cheu, Jain, Altman, Schoenholz, Toizer, Miserendino, Agarwal, Culver, Ethersmith, Gray, Grove, Metzger, Hermani, Jain, Zhao, Wu, Jomoto, Wu, Shuaiqi, Xia, Phene, Papay, Narayanan, Coffey, Lee, Hall, Balaji, Broda, Stramer, Xu, Gogineni,
  Christianson, Sanders, Patwardhan, Cunninghman, Degry, Dimson, Raoux, Shadwell, Zheng, Underwood, Markov, Sherbakov, Rubin, Stasi, Kaftan, Heywood, Peterson, Walters, Eloundou, Qi, Moeller, Monaco, Kuo, Fomenko, Chang, Zheng, Zhou, Manassra, Sheu, Zaremba, Patil, Qian, Kim, Cheng, Zhang, He, Zhang, Jin, Dai, and Malkov}]{gpt4omini}
OpenAI, Aaron Hurst, Adam Lerer, Adam~P. Goucher, Adam Perelman, Aditya Ramesh, Aidan Clark, AJ~Ostrow, Akila Welihinda, Alan Hayes, Alec Radford, Aleksander Mądry, Alex Baker-Whitcomb, Alex Beutel, Alex Borzunov, Alex Carney, Alex Chow, Alex Kirillov, Alex Nichol, and 400 others. 2024.
\newblock \href {https://arxiv.org/abs/2410.21276} {{GPT-4o System Card}}.
\newblock \emph{Preprint}, arXiv:2410.21276.

\bibitem[{Ouyang et~al.(2022)Ouyang, Wu, Jiang, Almeida, Wainwright, Mishkin, Zhang, Agarwal, Slama, Ray, Schulman, Hilton, Kelton, Miller, Simens, Askell, Welinder, Christiano, Leike, and Lowe}]{ouyang2022traininglanguagemodelsfollow}
Long Ouyang, Jeff Wu, Xu~Jiang, Diogo Almeida, Carroll~L. Wainwright, Pamela Mishkin, Chong Zhang, Sandhini Agarwal, Katarina Slama, Alex Ray, John Schulman, Jacob Hilton, Fraser Kelton, Luke Miller, Maddie Simens, Amanda Askell, Peter Welinder, Paul Christiano, Jan Leike, and Ryan Lowe. 2022.
\newblock \href {https://arxiv.org/abs/2203.02155} {Training language models to follow instructions with human feedback}.
\newblock \emph{Preprint}, arXiv:2203.02155.

\bibitem[{Qwen et~al.(2025)Qwen, Yang, Yang, Zhang, Hui, Zheng, Yu, Li, Liu, Huang, Wei, Lin, Yang, Tu, Zhang, Yang, Yang, Zhou, Lin, Dang, Lu, Bao, Yang, Yu, Li, Xue, Zhang, Zhu, Men, Lin, Li, Tang, Xia, Ren, Ren, Fan, Su, Zhang, Wan, Liu, Cui, Zhang, and Qiu}]{qwen2025qwen25technicalreport}
Qwen, An~Yang, Baosong Yang, Beichen Zhang, Binyuan Hui, Bo~Zheng, Bowen Yu, Chengyuan Li, Dayiheng Liu, Fei Huang, Haoran Wei, Huan Lin, Jian Yang, Jianhong Tu, Jianwei Zhang, Jianxin Yang, Jiaxi Yang, Jingren Zhou, Junyang Lin, and 24 others. 2025.
\newblock \href {https://arxiv.org/abs/2412.15115} {{Qwen2.5 Technical Report}}.
\newblock \emph{Preprint}, arXiv:2412.15115.

\bibitem[{Rodr{\'i}guez-Cantelar et~al.(2023)Rodr{\'i}guez-Cantelar, Zhang, Tang, Shi, Ghazarian, Sedoc, Fernando~D{'}Haro, and Rudnicky}]{rodriguez-cantelar-etal-2023-overview}
Mario Rodr{\'i}guez-Cantelar, Chen Zhang, Chengguang Tang, Ke~Shi, Sarik Ghazarian, Jo{\~a}o Sedoc, Luis Fernando~D{'}Haro, and Alexander~I. Rudnicky. 2023.
\newblock \href {https://aclanthology.org/2023.dstc-1.28/} {Overview of robust and multilingual automatic evaluation metricsfor open-domain dialogue systems at {DSTC} 11 track 4}.
\newblock In \emph{Proceedings of The Eleventh Dialog System Technology Challenge}, pages 260--273, Prague, Czech Republic. Association for Computational Linguistics.

\bibitem[{Shuster et~al.(2022)Shuster, Xu, Komeili, Ju, Smith, Roller, Ung, Chen, Arora, Lane, Behrooz, Ngan, Poff, Goyal, Szlam, Boureau, Kambadur, and Weston}]{shuster2022blenderbot3deployedconversational}
Kurt Shuster, Jing Xu, Mojtaba Komeili, Da~Ju, Eric~Michael Smith, Stephen Roller, Megan Ung, Moya Chen, Kushal Arora, Joshua Lane, Morteza Behrooz, William Ngan, Spencer Poff, Naman Goyal, Arthur Szlam, Y-Lan Boureau, Melanie Kambadur, and Jason Weston. 2022.
\newblock \href {https://arxiv.org/abs/2208.03188} {Blenderbot 3: a deployed conversational agent that continually learns to responsibly engage}.
\newblock \emph{Preprint}, arXiv:2208.03188.

\bibitem[{Vidgen et~al.(2024)Vidgen, Agrawal, Ahmed, Akinwande, Al-Nuaimi, Alfaraj, Alhajjar, Aroyo, Bavalatti, Bartolo, Blili-Hamelin, Bollacker, Bomassani, Boston, Campos, Chakra, Chen, Coleman, Coudert, Derczynski, Dutta, Eisenberg, Ezick, Frase, Fuller, Gandikota, Gangavarapu, Gangavarapu, Gealy, Ghosh, Goel, Gohar, Goswami, Hale, Hutiri, Imperial, Jandial, Judd, Juefei-Xu, Khomh, Kailkhura, Kirk, Klyman, Knotz, Kuchnik, Kumar, Kumar, Lengerich, Li, Liao, Long, Lu, Luger, Mai, Mammen, Manyeki, McGregor, Mehta, Mohammed, Moss, Nachman, Naganna, Nikanjam, Nushi, Oala, Orr, Parrish, Patlak, Pietri, Poursabzi-Sangdeh, Presani, Puletti, Röttger, Sahay, Santos, Scherrer, Sebag, Schramowski, Shahbazi, Sharma, Shen, Sistla, Tang, Testuggine, Thangarasa, Watkins, Weiss, Welty, Wilbers, Williams, Wu, Yadav, Yang, Zeng, Zhang, Zhdanov, Zhu, Liang, Mattson, and Vanschoren}]{vidgen2024introducingv05aisafety}
Bertie Vidgen, Adarsh Agrawal, Ahmed~M. Ahmed, Victor Akinwande, Namir Al-Nuaimi, Najla Alfaraj, Elie Alhajjar, Lora Aroyo, Trupti Bavalatti, Max Bartolo, Borhane Blili-Hamelin, Kurt Bollacker, Rishi Bomassani, Marisa~Ferrara Boston, Siméon Campos, Kal Chakra, Canyu Chen, Cody Coleman, Zacharie~Delpierre Coudert, and 81 others. 2024.
\newblock \href {https://arxiv.org/abs/2404.12241} {Introducing v0.5 of the ai safety benchmark from mlcommons}.
\newblock \emph{Preprint}, arXiv:2404.12241.

\bibitem[{Xu et~al.(2021)Xu, Ju, Li, Boureau, Weston, and Dinan}]{xu-etal-2021-bot}
Jing Xu, Da~Ju, Margaret Li, Y-Lan Boureau, Jason Weston, and Emily Dinan. 2021.
\newblock \href {https://doi.org/10.18653/v1/2021.naacl-main.235} {Bot-adversarial dialogue for safe conversational agents}.
\newblock In \emph{Proceedings of the 2021 Conference of the North American Chapter of the Association for Computational Linguistics: Human Language Technologies}, pages 2950--2968, Online. Association for Computational Linguistics.

\bibitem[{Yin et~al.(2024)Yin, Qiu, Huang, Chang, and Peng}]{yin2024safeworld}
Da~Yin, Haoyi Qiu, Kung-Hsiang Huang, Kai-Wei Chang, and Nanyun Peng. 2024.
\newblock \href {https://openreview.net/forum?id=VZQmIoDGBG} {Safeworld: Geo-diverse safety alignment}.
\newblock In \emph{The Thirty-eighth Annual Conference on Neural Information Processing Systems}.

\bibitem[{Yu et~al.(2024)Yu, Li, Liao, Wang, Zuchen, Mi, and Hong}]{yu-etal-2024-cosafe}
Erxin Yu, Jing Li, Ming Liao, Siqi Wang, Gao Zuchen, Fei Mi, and Lanqing Hong. 2024.
\newblock \href {https://doi.org/10.18653/v1/2024.emnlp-main.968} {{C}o{S}afe: Evaluating large language model safety in multi-turn dialogue coreference}.
\newblock In \emph{Proceedings of the 2024 Conference on Empirical Methods in Natural Language Processing}, pages 17494--17508, Miami, Florida, USA. Association for Computational Linguistics.

\bibitem[{Zhang et~al.(2021)Zhang, Sedoc, D'Haro, Banchs, and Rudnicky}]{zhang2011automatic}
Chen Zhang, João Sedoc, Luis~Fernando D'Haro, Rafael Banchs, and Alexander Rudnicky. 2021.
\newblock \href {https://arxiv.org/abs/2111.02110} {Automatic evaluation and moderation of open-domain dialogue systems}.
\newblock \emph{Preprint}, arXiv:2111.02110.

\bibitem[{Zhang et~al.(2024)Zhang, Sedoc, and Levina}]{zhang2024back}
Lining Zhang, Jo{\~a}o Sedoc, and Natalia Levina. 2024.
\newblock \href {https://aisel.aisnet.org/icis2024/adv_theory/adv_theory/4} {Back to principles: Theory-driven evaluation of ai-based conversational agents}.
\newblock In \emph{Forty-Fifth International Conference on Information Systems}.

\end{thebibliography}

\end{document}